\documentclass{article}
\usepackage{spconf,amsmath,graphicx}
\usepackage{booktabs}
\usepackage{multirow}
\usepackage{pifont}
\usepackage{url}
\newcommand{\cmark}{\ding{51}}

\title{A BI-PYRAMID MULTIMODAL FUSION METHOD FOR THE diagnosis \\ OF BIPOLAR DISORDERS}
\name{Guoxin Wang$^{1,*}$, Sheng Shi$^{2,*}$, Shan An$^{3}$, Fengmei Fan$^{4}$, Wenshu Ge$^{3}$, Qi Wang$^{2,\dag}$, Feng Yu$^{1,\dag}$, Zhiren Wang$^{4}$ \thanks{$^{\dag}$Corresponding authors}
\thanks{$^{*}$Equal contribution}}

\address{$^{1}$College of Biomedical Engineering \& Instrument Science, Zhejiang University, Hangzhou, China, \\
$^{2}$College of Sciences, Northeastern University, Shenyang, China, \\
$^{3}$JD Health International Inc., Beijing, China, \\
$^{4}$Beijing Huilongguan Hospital, Peking University Huilongguan Clinical Medical School, Beijing, China.}
\begin{document}
%
\maketitle
\begin{abstract}
Previous research on the diagnosis of Bipolar disorder has mainly focused on resting-state functional magnetic resonance imaging. However, their accuracy can not meet the requirements of clinical diagnosis. Efficient multimodal fusion strategies have great potential for applications in multimodal data and can further improve the performance of medical diagnosis models. In this work, we utilize both sMRI and fMRI data and propose a novel multimodal diagnosis model for bipolar disorder. The proposed Patch Pyramid Feature Extraction Module extracts sMRI features, and the spatio-temporal pyramid structure extracts the fMRI features. Finally, they are fused by a fusion module to output diagnosis results with a classifier. Extensive experiments show that our proposed method outperforms others in balanced accuracy from 0.657 to 0.732 on the OpenfMRI dataset, and achieves the state of the art.
\end{abstract}
\section{Introduction}
Bipolar disorder (BD) is a severe mood disorder, typically defined by alternating episodes of two emotional states: depressive symptoms and manic symptoms. The diagnosis of BD relies on subjective reports from patients and clinical observations. Making the diagnosis and treatment is still challenging. Although the pathophysiology of BD is not clear, the application of functional magnetic resonance imaging (fMRI) and structural magnetic resonance imaging (sMRI) techniques has helped us better understand the changes in brain function and structure in BD patients\cite{Waller2021ReviewingAO, Li2020IdentificationOB}.  

Applying fMRI and sMRI techniques can help us understand the neurobiological mechanisms of BD in-depth, providing more objective and accurate diagnostic methods. In this study, our objective is to introduce a pioneering multimodal diagnosis model for BD, leveraging both sMRI and fMRI data. We aim to achieve a model accuracy that aligns with the demands of clinical diagnosis. The proposed Patch Pyramid Feature Extraction Module (P2FEM) extracts sMRI features, and the Spatio-temporal Feature Aggregation Module (SFAM) extracts fMRI features. These features are subsequently fused by a fusion module to yield diagnosis results through a classifier. Extensive experiments have shown that the proposed method has achieved advanced results in BD disease diagnosis. The contributions of this paper are as follows: 
\begin{itemize}
    \item An end-to-end BD diagnostic framework called the bi-Pyramid Multimodal Fusion Method is proposed. It can accurately and efficiently extract information from sMRI and fMRI data and make diagnoses.
    \item Extensive experiments show that our proposed method outperforms others in balanced accuracy from 0.657 to 0.732 on the OpenfMRI dataset, and achieves the state of the art. 
\end{itemize}

\begin{figure*}[th] 
  \centering
  \includegraphics[width=\linewidth]{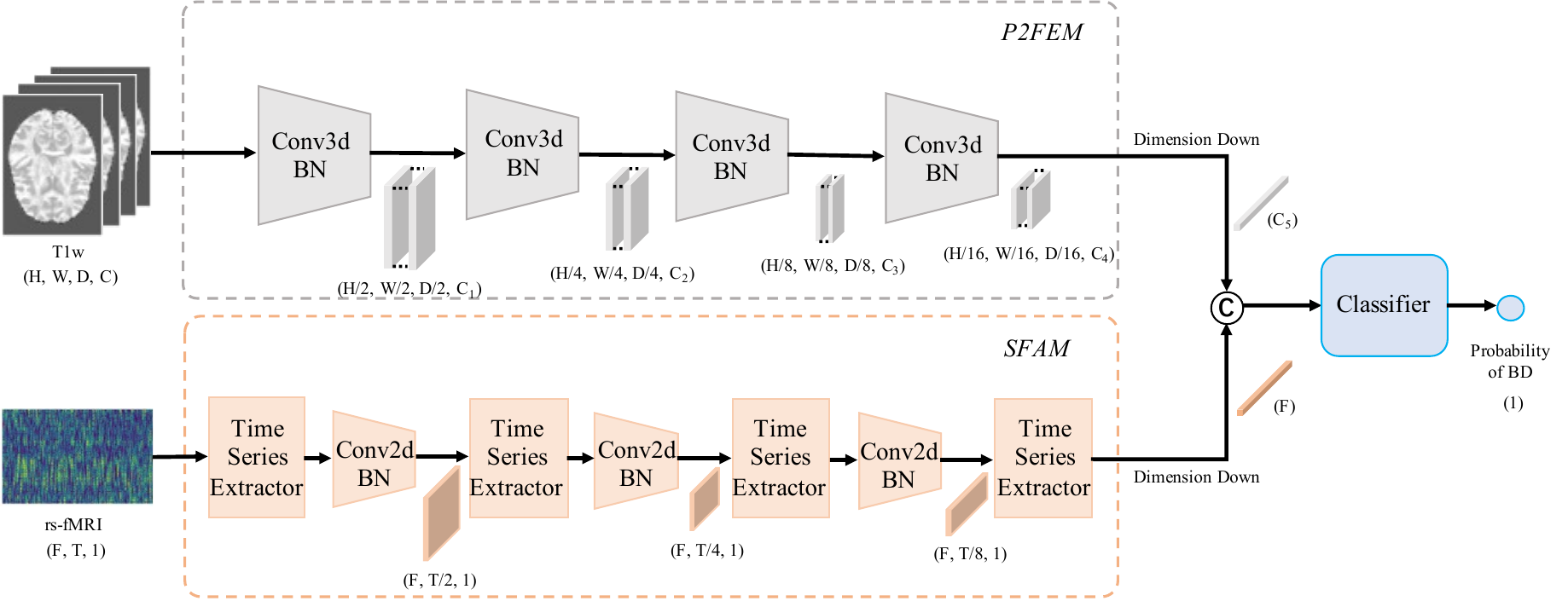}
  \caption{An overview of the proposed method. Rs-fMRI and T1w are respectively fed into separate encoder branches for feature extraction. The Patch Pyramid Feature Extraction Module consists of four consecutive convolutional layers, while the Spatio-temporal Feature Aggregation Module comprises concatenated spatial feature extraction modules and temporal feature extraction modules. After dimensional reduction, the extracted features are concatenated and inputted into a classifier for prediction and output of the prediction results.}
\label{fig:over}
\end{figure*}

\section{Related Work}
\label{sec:related}

sMRI data refers to structural imaging data of the patient's brain. \cite{Dai2022FirstGD} designed two Attention modules, which combine local features and global features through dual-branch learning and apply them to 3D fMRI classification. \cite{Xing2022NestedFormerNM, Zhang2022mmFormerMM} conduct research based on multimodal data, and use Transformers to learn the features between modalities. \cite{9852280} proposes a novel attention-based 3D Multi-scale CNN model to capture multiple spatial-scale features. \cite{LI2021101882} designed a convolutional neural network framework based on sMRI for automatic diagnosis. Compared with transformer-based methods, convolution-based methods have lower computational complexity.

fMRI data refer to continuous images of the patient over a period. \cite{Lu2016DiscriminativeAO} analyzed the gray matter images of schizophrenia MRI using VBM and ROI and classified schizophrenia patients using SVM. \cite{Jeon2020EnrichedRL} directly input the time sequence features of ROI regions into the network to learn spatial and temporal features using convolution and attention mechanisms. \cite{Wang2021ModelingDC} uses an overlapping sliding window to divide rs-fMRI time series into segments, then builds a longitudinally ordered functional connectivity network. \cite{kanyal2023multi} applied XGBoost to diagnose schizophrenia based on sMRI, fMRI, and SNPs signals. Additionally, \cite{ghayem2023new} employed dictionary learning and independent component analysis for schizophrenia diagnosis using fMRI data.

\section{Methods}

Based on sMRI and fMRI data, a convolution-based model is proposed. Figure~\ref{fig:over} shows the overall structure. Two encoder branches are designed to construct the rs-fMRI and the T1w, respectively.

\subsection{Patch Pyramid Feature Extraction Module}
To reduce the complexity, a novel Patch Pyramid Feature Extraction Module (P2FEM) based on convolution is proposed. Existing convolutional neural networks consist of convolutional layers, pooling layers, and BN layers. The pooling layer is used to reduce the number of model parameters, increase the receptive field, alleviate overfitting, and so on. We suppose that the operation convolution and pooling have some redundancy. 

To improve the model's efficiency, we removed the pooling layer and adjusted the convolutional layer. We use a large convolutional kernel to expand the model's receptive field and improve its learning ability. To alleviate overfitting, we set the stride to reduce the density of convolution. The down-sample procedure is completed at the same time. In addition, we use a group strategy to further improve efficiency.

\begin{figure}[h]
\centerline{\includegraphics[width=\columnwidth]{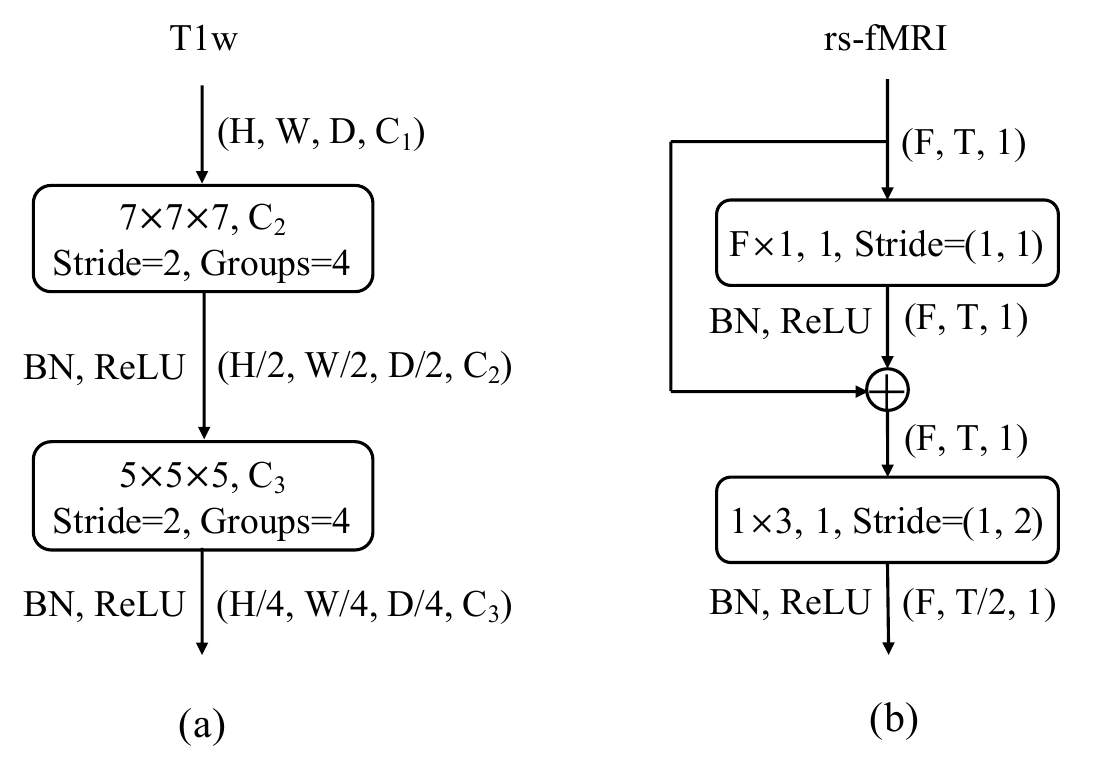}}
\caption{The structure of P2FEM and SFAM. (a) The structure of P2FEM. (b) The structure of SFAM.}
\label{fig2}
\end{figure}

Figure~\ref{fig2}(a) illustrates the specific structure of P2FEM. Compared with traditional convolutional neural networks, P2FEM has fewer parameters and higher efficiency. And the extracted features have stronger generalization ability. 

\subsection{Spatio-temporal Feature Aggregation Module}
FMRI data is a continuous stream with a huge size. Direct processing of fMRI will bring a huge computational cost. Existing methods typically first extract a specific number of brain regions, and then extract functional connectivity matrices from that. To construct an efficient BD diagnosis network, this paper also analyzes the features of brain regions with the connectivity matrices. Assuming that the preprocessed fMRI data $ P=\{(x_{i1},\cdots,x_{iN})|i = 1,\cdots,M\} $, where M represents the number of frames in fMRI data, and N represents the number of brain regions. A spatio-temporal feature aggregation module(SFAM) is proposed to learn the temporal and spatial features of fMRI data. Figure~\ref{fig2}(b) explains its specific structure.

We use concatenated spatial and temporal feature extraction modules as a spatio-temporal feature extraction unit. The spatial feature extraction module consists of fully connected layers and BN layers. The fully connected layer can learn the global features of fMRI data. In addition, we use residual structures to further improve the accuracy of the model. As the depth of the network increases, we keep the number of brain regions N unchanged to reduce the complexity and avoid redundant features. For the temporal dimension, we use convolutional layers to extract temporal features. As shown in Figure~\ref{fig2}(b), the features between adjacent frames are aggregated through a convolution function. In addition, we downsample the fMRI data and expand the model's receptive field in the temporal dimension.

In the last, a fusion module is proposed to fuse the features of the two modalities. Firstly, the sMRI data is processed into the target size using the flatten function and fully connected neural network. Then, it is concatenated with fMRI data with the concat function. Finally, a classifier is used to process the features and obtain the predicted probability of BD.

\section{Experiment}
To validate the effectiveness of our proposed method, we conducted experiments on two different datasets, including a publicly available dataset, OpenfMRI, and our collected dataset. Multiple evaluation metrics were used to compare the performance of different methods, including BACC, F1 score, sensitivity, and specificity. We designed two binary classification experiments (BD vs HC) for comparison.

\subsection{Participants}
Our collected dataset used in this study was collected from Beijing Huilongguan Hospital between 2014 and 2018\footnote{This study was performed in line with the principles of the Declaration of Helsinki. Approval was granted by the Ethics Committee of Beijing HuiLongGuan Hospital.}. It consists of 91 samples, including 39 healthy controls (HC, 22 males and 17 females, average age = $31.9\pm6.5$ years) and 52 patients with bipolar disorder (BD, 33 males and 19 females, average age = $27.1\pm6.2$ years), as shown in Table~\ref{tab:participant}. The patients were diagnosed with BD using the Diagnostic and Statistical Manual for Mental Disorders-Fourth Edition (DSM-IV) in the Psychiatric Hospital of Zhumadian in Henan Province, China. Each sample data contains two modalities of data: T1-weighted Anatomical MPRAGE and resting-state fMRI. The size of the T1w data is 256256188, and the rs-fMRI data contains 210 frames of 646433 voxels.

\begin{table}[ht]
\begin{center}
\caption{Demographics and clinical characteristics of participants. BD = Bipolar disorder; HAMD = 17-items Hamilton depression scale; HAMA = Hamilton anxiety scale; YMRS = Young Mania Rating Scale.}
\label{tab:participant}
\resizebox{0.48\textwidth}{!}{
    \begin{tabular}{ccccc}
        \toprule  
        \multicolumn{1}{l}{\multirow{2}*{}} & \multicolumn{1}{l}{\multirow{2}*{BD(n=52)}} & \multicolumn{1}{l}{\multirow{2}*{Controls(n=39)}} & \multicolumn{2}{c}{BD vs. Controls}  \\
        \cmidrule(r){4-5}
        \multicolumn{1}{l}{} & \multicolumn{1}{l}{} & \multicolumn{1}{l}{} & $\chi^{2/t}$ & $P$ \\
        \midrule
        Sex(M/F) & 33/19 & 22/17 & 0.24 & 0.62 \\
        Age(years) & 27.1$\pm$6.2 & 31.9$\pm$6.5 & 3.57 & 0.001 \\
        Illness duration(years) & 5.3$\pm$3.2 \\
        Onset age(years) & 22.6$\pm$6.8 \\
        Number of manic episodes & 2.6$\pm$1.6 \\
        Number of depression episodes & 1.5$\pm$1.2 \\
        HAMD score & 9.4$\pm$9.6 \\
        HAMA score & 5.7$\pm$7.0 \\
        YMRS score & 22.5$\pm$12.4 \\
        Mood Stabilizers n(\%) & 40(83.3\%) \\
        Antidepressants n(\%) & 12(25\%) \\
        Antipsychotics n(\%) & 33(68.8\%) \\
        \bottomrule
    \end{tabular}}
\end{center}
\end{table}

To further validate the effectiveness of our proposed method, we also conducted experiments using the publicly available dataset OpenfMRI. OpenfMRI is a commonly used public dataset (\url{https://openfmri.org/dataset/ds000030/}, Revision: 1.0.5). We selected healthy controls and patients with BD from this dataset to maintain consistency with our collected dataset. After filtering, a total of 171 samples were obtained (122 HC and 49 patients with BD), and only T1-weighted Anatomical MPRAGE and Resting State fMRI data were used for training and testing.

\subsection{Experiment Setup}
We verified the effectiveness of the proposed method based on both our collected dataset and the public dataset OpenfMRI with five-fold cross-validation. A Tesla P40 (24G) graphics card was used to train the network. To evaluate the performance of the proposed method, four metrics were used in the experiments: balanced accuracy (BACC), f1 score, sensitivity (SEN), and specificity (SPEC). For imbalanced data, BACC is more representative than accuracy. The formulas for calculating are as follows:

\begin{equation}F1 = \frac{2*TP}{2*TP+FP+FN}.\label{F1}\end{equation}

\begin{equation}SEN = \frac{TP}{TP+FN}.\label{SEN}\end{equation}

\begin{equation}SPEC = \frac{TN}{TN+FP}.\label{SPEC}\end{equation}

\begin{equation}BACC = \frac{SEC+SPEC}{2}.\label{BACC}\end{equation} where TP (True positives) represents samples that are predicted as positive and actually positive, TN (True negatives) represents samples that are predicted as negative and actually negative, FP (False positives) represents samples that are predicted as positive but actually negative, and FN (False negatives) represents samples that are predicted as negative but actually positive.

\section{Results}

\subsection{Comparison with existing Methods}
To validate the effectiveness of the proposed method, we applied it to both the public dataset OpenfMRI and our collected dataset. The bold numbers indicate the top two performances in each evaluation metric. To be fair, except for the network architecture, all methods used the same experimental settings.

Table~\ref{tab:pr} shows the classification results of different methods on our collected BD dataset. Overall, our proposed method significantly outperforms existing methods in both single and multimodal fusion schemes, achieving 0.736 and 0.766 BACC on two different classifiers, respectively. Meanwhile, our method demonstrates a significant improvement in the F1 score, indicating good predictive performance in terms of precision and recall. In addition, our method achieved a good balance between sensitivity and specificity (0.680 vs 0.793, 0.789 vs 0.744), effectively reducing the rates of missed diagnosis and misdiagnosis for BD.

\begin{table}[ht]
\begin{center}
\caption{The results on our collected dataset. The top-two results are highlighted in bold.}
\label{tab:pr}
\resizebox{0.45\textwidth}{!}{
\begin{tabular}{ccccc}
\toprule
Method      & BACC           & F1             & SEN            & SPEC           \\ \midrule
PCC$+$SVM\cite{Yan2019DiscriminatingSU}     & 0.570          & 0.595          & 0.621          & 0.519          \\
PCC$+$MLP\cite{Yan2019DiscriminatingSU}     & 0.595          & 0.547          & 0.545          & 0.646          \\
Late Fusion\cite{Narazani2022IsAP} & 0.686          & 0.542          & \textbf{0.774}          & 0.597          \\
Ours-Dense        & \textbf{0.736} & \textbf{0.708} & 0.680 & \textbf{0.793} \\
Ours-Linear        & \textbf{0.766} & \textbf{0.796} & \textbf{0.789} & \textbf{0.744} \\ \bottomrule
\end{tabular}}    
\end{center}
\end{table}

We further validate the effectiveness of our approach on OpenfMRI. Table~\ref{tab:op} shows the classification results. The experimental results of LSTM, DGM, and sw-DGM are taken from the paper~\cite{Matsubara2018StructuredDG}. To ensure a fair comparison, we use parameters that are as similar as possible to those of these methods. Among all the compared methods, Ours-Linear shows a significant improvement in both BACC and F1 score, increasing by 11.4$\%$ and 23.0$\%$ respectively. The results demonstrate that our proposed method has a significantly effective performance in detecting bipolar disorder.

\begin{table}[ht]
\begin{center}
\caption{The results on the public dataset OpenfMRI. The top-two results are highlighted in bold.}
\label{tab:op}
\resizebox{0.45\textwidth}{!}{
\begin{tabular}{ccccc}
\toprule
Method      & BACC           & F1             & SEN            & SPEC           \\ \midrule
PCC$+$SVM\cite{Yan2019DiscriminatingSU}     & 0.555          & 0.403          & 0.422          & 0.689          \\
PCC$+$MLP\cite{Yan2019DiscriminatingSU}     & 0.605          & 0.323          & 0.239          & \textbf{0.963} \\
LSTM\cite{Dvornek2017IdentifyingAF}        & 0.571          & -              & \textbf{0.802}          & 0.340          \\
DGM\cite{tashiro2017deep}         & 0.619          & -              & 0.650          & 0.587          \\
sw-DGM\cite{Matsubara2018StructuredDG}      & 0.622          & -              & \textbf{0.844} & 0.401          \\
Late Fusion\cite{Narazani2022IsAP} & 0.657          & 0.496          & 0.767          & 0.548          \\
Ours-Dense        & \textbf{0.685} & \textbf{0.553} & 0.650         & 0.720   \\
Ours-Linear        & \textbf{0.732} & \textbf{0.610} & 0.735          & \textbf{0.730}          \\  \bottomrule
\end{tabular}}    
\end{center}
\end{table}

\subsection{Ablation Study}
To validate the necessity of multimodal fusion, we also conducted experiments using a linear classifier network on both our collected and OpenfMRI dataset with three settings: using a single modality of sMRI or fMRI branch for diagnosis and using both sMRI and fMRI after feature fusion for diagnosis. All experimental settings, except for the input modality, were kept the same. According to Table~\ref{tab:ablation}, the model that combines two modalities performs better than the model that uses only a single modality, which indicates the effectiveness of the multimodal fusion mechanism. 

\begin{table}[ht]
\begin{center}
\caption{Ablation study of our methods using a linear classifier net on our collected dataset and the public dataset OpenfMRI. The best results are highlighted in bold.}
\label{tab:ablation}
\resizebox{0.48\textwidth}{!}{
\begin{tabular}{ccc|cccc}
\toprule
Datasets                          & sMRI     & fMRI     & BACC           & F1             & SEN            & SPEC           \\ \midrule
\multirow{3}{*}{Our collected dataset}  & \cmark &                       & 0.718          & 0.762          & 0.769          & \textbf{0.667} \\
                                  &                       & \cmark & 0.694          & 0.723          & 0.743          & 0.646          \\
                                  & \cmark & \cmark & \textbf{0.776} & \textbf{0.829} & \textbf{0.885} & \textbf{0.667} \\ \midrule
\multirow{3}{*}{OpenfMRI dataset} & \cmark &                       & 0.575          & 0.257          & 0.259          & \textbf{0.890} \\
                                  &                       & \cmark & 0.658          & 0.474          & 0.496          & 0.821          \\
                                  & \cmark & \cmark & \textbf{0.739} & \textbf{0.599} & \textbf{0.744} & 0.733  \\
  
  \bottomrule
\end{tabular}
}
\end{center}
\end{table}

\section{Conclusion}
In this paper, we propose a novel bi-pyramid multimodal fusion algorithm (BPM-Fusion) to provide a new perspective on disease diagnosis. The model takes sMRI and fMRI data as inputs. Extensive experiments show that our proposed method outperforms others in balanced accuracy from 0.657 to 0.732 on the OpenfMRI dataset, and achieves the state of the art. In future work, we will explore different fusion methods to further develop the application of multimodal data in disease diagnosis.

\vfill\pagebreak

\bibliographystyle{IEEEbib}
\bibliography{strings,refs}

\end{document}